\documentclass[runningheads]{llncs}
\usepackage[T1]{fontenc}
\usepackage{orcidlink}
\usepackage{graphicx}
\usepackage{amsmath,amssymb}
\usepackage{booktabs}
\usepackage{multirow}
\usepackage{tabularx}
\usepackage{marvosym}
\usepackage[table]{xcolor}
\newcommand{\hl}[1]{\cellcolor{gray!12}#1}
\usepackage[normalem]{ulem}
\useunder{\uline}{\ul}{}
\usepackage{url}

\newcommand{\softmax}{\mathrm{softmax}}

\begin{document}

\title{Region-Graph Optimal Transport Routing for Mixture-of-Experts Whole-Slide Image Classification}
\titlerunning{ROAM: Region-Graph Optimal Transport Routing for WSI Classification}

\author{
Xin Tian\inst{1}\orcidlink{0000-0003-1168-5298}$^{(\textrm{\Letter})}$ \and
Jiuliu Lu\inst{2} \and
Ephraim Tsalik\inst{3} \and
Bart Wanders\inst{2} \and
Colleen Knoth\inst{3} \and
Julian Knight\inst{1}
}

\authorrunning{X. Tian et al.}

\institute{
Centre for Human Genetics, University of Oxford, Oxford, UK\\
\email{xin.tian@well.ox.ac.uk, julian.knight@well.ox.ac.uk}
\and
Beckman Coulter Diagnostics, Brea, CA, USA
\and
Danaher Corporation, Washington, DC, USA
}


\maketitle

\begin{abstract}

Multiple Instance Learning (MIL) is the dominant framework for gigapixel
whole-slide image (WSI) classification in computational pathology. However, current MIL aggregators route all instances through a shared pathway, constraining their capacity to specialise across the pathological heterogeneity inherent in each slide. Mixture-of-Experts (MoE) methods offer a natural remedy by partitioning instances across specialised expert subnetworks; yet unconstrained softmax routing may yield highly imbalanced utilisation, where one or a few experts absorb most routing mass, collapsing the mixture back to a near-single-pathway solution. 
To address these limitations, we propose ROAM (\textbf{R}egion-graph \textbf{O}ptim\textbf{A}l-transport \textbf{M}ixture-of-experts), a spatially aware MoE-MIL aggregator that routes region tokens to expert poolers via capacity-constrained entropic optimal transport, promoting balanced expert utilisation by construction. ROAM operates on spatial region tokens—obtained by compressing dense patch bags into spatially binned units that align routing with local tissue neighbourhoods—and introduces two key mechanisms: (i) region-to-expert assignment formulated as entropic optimal transport (Sinkhorn) with explicit per-slide capacity marginals, enforcing balanced expert utilisation without auxiliary load-balancing losses; and (ii) graph-regularised Sinkhorn iterations that diffuse routing assignments over the spatial region graph, encouraging neighbouring regions to coherently route to the same experts.
Evaluated on four WSI benchmarks with frozen foundation-model patch embeddings, ROAM achieves performance competitive against strong MIL and MoE baselines, and on NSCLC generalisation (TCGA$\rightarrow$CPTAC) reaches external AUC $0.845\pm0.019$. The code will be made available upon publication.

\keywords{Multiple instance learning \and computational pathology \and
mixture of experts \and optimal transport \and whole-slide image}
\end{abstract}

\section{Introduction}

Whole-slide images (WSIs) contain gigapixel-scale tissue with extreme morphological and spatial heterogeneity, where diagnostically relevant evidence is often sparse and locally clustered. With the rise of pathology foundation models, it has become standard to represent a WSI as a bag of frozen patch embeddings~\cite{chen2024uni,gigapath} and learn a slide-level predictor under weak supervision. 
Multiple Instance Learning (MIL) is the dominant framework for this setting~\cite{ilse2018abmil,campanella2019clinical}: treat each patch embedding as an instance, aggregate across the bag (one bag
per slide), and predict a slide-level label with only bag-level supervision. Typically, gated attention (ABMIL~\cite{ilse2018abmil}) learns instance-level importance weights. Instance-level clustering (CLAM~\cite{lu2021clam}) partitions instances into subgroups before pooling. Dual-stream architectures (DSMIL~\cite{li2021dsmil}) combine max-instance and attention pathways. Transformer-based aggregators (TransMIL~\cite{shao2021transmil}) model pairwise instance correlations, while low-rank approximations (ILRA~\cite{xiang2023ilra}) improve scalability. However, all these methods share a structural limitation: every instance is processed through a single shared pathway, forcing morphologically and semantically distinct populations into a common parameter space. This conflation limits both specialisation and interpretability.

Mixture-of-Experts (MoE) methods offer a principled alternative by decomposing the shared pathway into specialised subnetworks. MAMMOTH~\cite{shao2026mammoth} replaces linear projections in MIL aggregators with factorised mini-expert modules. PAMoE~\cite{wu2025pamoe} uses expert choice routing supervised by tissue prototypes. Graph of Tokens~\cite{nguyen2025graphoftokens} proposes inter-token similarity-aware routing for sparse MoE in vision. In a complementary direction, OTSurv~\cite{ren2025otsurv} applies semi-relaxed Sinkhorn optimal transport~\cite{cuturi2013sinkhorn,knight2008sinkhorn} to learnable prototypes for survival prediction, showing the benefit of capacity-constrained assignment over standard attention pooling. 

Despite these advances, two challenges remain underaddressed. First, unconstrained softmax dispatch in MoE imposes no constraint on aggregate routing mass: if one expert's gating score dominates, it absorbs the majority of instances, collapsing the MoE into a near-single-pathway model~\cite{fedus2022switch}. Auxiliary load-balancing losses partially mitigate this but introduce a sensitive hyperparameter and provide no hard guarantee~\cite{wang2024auxiliary,zhou2022expertchoice}. Second, patch-wise routing is typically computed per instance with limited spatial context, despite the fact that pathologically meaningful structures in WSIs are spatially coherent~\cite{fang2024sammil}. This can fragment neighbouring coherent patches, thus weakening interpretability and destabilising expert learning.

To address these challenges, we propose ROAM (\textbf{R}egion-graph \textbf{O}ptim\textbf{A}l-transport \textbf{M}ixture-of-experts), a spatially aware MoE-MIL aggregator that routes spatial region tokens to expert poolers via capacity-constrained entropic optimal transport. 
First, ROAM compresses dense patch bags into spatial region tokens via grid binning, so routing operates on local tissue neighbourhoods rather than isolated patches. This makes routing tractable ($M{\ll}N$) and better aligns assignments with spatially coherent morphology. 
Second, region-to-expert assignment is formulated as an entropic optimal transport problem~\cite{khamis2024scalable,villani2009optimal} with explicit per-slide capacity marginals. Unlike softmax dispatch, which provides no utilisation control and is prone to routing imbalance~\cite{fedus2022switch}, Sinkhorn enforces balanced expert load in the routing plan by construction, avoiding auxiliary load-balancing losses~\cite{wang2024auxiliary}. 
Third, we interleave a lightweight graph smoothing step within unrolled Sinkhorn updates on a spatial region graph, encouraging spatially coherent routing distributions and reducing fragmentation induced by instance-independent routing~\cite{fang2024sammil}. Across four WSI benchmarks with frozen foundation-model embeddings, ROAM is competitive with strong MIL and MoE baselines, and on NSCLC generalisation (TCGA$\rightarrow$CPTAC) reaches external AUC $0.845\pm0.019$.
 
\section{Methodology}
\label{sec:method}
\subsection{Problem Formulation and Overview}
\begin{figure}[t]
\centering
\includegraphics[width=0.98\textwidth]{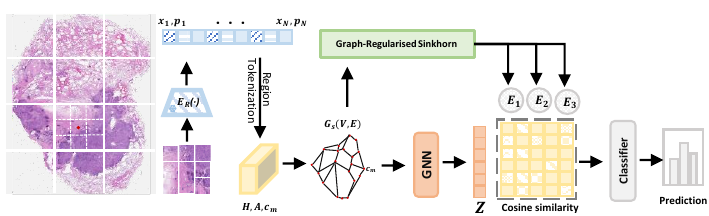}
\caption{Frozen patch embeddings are pooled into $M$ spatial region tokens.
A routing GNN on the region graph parameterises region-to-expert costs, and Sinkhorn optimal transport routes region mass to $E$ experts under per-slide capacity constraints (with optional graph smoothing).
Each expert performs gated-attention pooling over its routed regions, and expert outputs are fused for slide-level prediction.}
\label{fig:method}
\end{figure}

A WSI is represented as a bag of $N$ frozen patch embeddings and 2D coordinates
$\{(\mathbf{x}_i, \mathbf{p}_i)\}_{i=1}^{N}$, where
$\mathbf{x}_i\in\mathbb{R}^{d_{\mathrm{in}}}$ and
$\mathbf{p}_i\in\mathbb{R}^{2}$.
Given a slide-level label $y$, we predict logits $\hat{\mathbf{y}}$ by aggregating instance evidence with a MoE MIL aggregator. ROAM replaces patch-wise softmax dispatch with \emph{capacity-constrained} routing on \emph{spatial region tokens}, followed by per-expert attention pooling and expert fusion. Fig.~\ref{fig:method} summarises ROAM: region tokenisation $\rightarrow$ (graph-aware) capacity-constrained Sinkhorn routing $\rightarrow$ per-expert gated-attention pooling $\rightarrow$ expert fusion.

\subsection{Region tokenisation}
We first project patch embeddings to a working dimension:
$\mathbf{h}_i=\phi(\mathbf{x}_i)$, where
$\phi:\mathbb{R}^{d_{\mathrm{in}}}\rightarrow\mathbb{R}^{d}$
is a lightweight MLP (linear layer, nonlinearity, and dropout).

To make routing tractable and spatially local, we compress the patch bag into $M{\ll}N$ \emph{region tokens} using grid binning in normalised coordinate space.
Let $\rho(i)\in\{1,\dots,M\}$ denote the region assignment of patch $i$ and $R_m=\{i:\rho(i)=m\}$ the set of patches in region $m$.
We compute region features, masses, and centroids:
\begin{equation}
H^{0}_m=\frac{1}{|R_m|}\sum_{i\in R_m} h_i,\quad
A_m=|R_m|,\quad
\mathbf{c}_m=\frac{1}{|R_m|}\sum_{i\in R_m} \mathbf{p}_i.
\end{equation}
Stacking over regions yields $H^{0}\in\mathbb{R}^{M\times d}$, $A\in\mathbb{R}^{M}$, and $\mathbf{P}\in\mathbb{R}^{M\times 2}$.
Region tokenisation reduces the routing computation from $O(NE)$ to $O(ME)$ and provides routing units that correspond to local tissue neighbourhoods, which supports the spatial-coherence regularisation introduced in \S\ref{sec:graph_sink}.

\subsection{Capacity-Constrained Optimal Transport Routing}
\label{sec:ot}

We assign $M$ region tokens to $E$ experts via entropic optimal
transport with per-slide capacity marginals.
Let $\Pi\in\mathbb{R}_{+}^{M\times E}$ denote the routing plan, where
$\Pi_{m,e}$ is the mass routed from region $m$ to expert $e$.

\subsubsection{Routing context via spatial region graph.}
We build a $k_{\mathrm{NN}}$-nearest-neighbour graph
$\mathcal{G}_s=(\mathcal{V},\mathcal{E})$ over region centroids
$\{\mathbf{c}_m\}$.
A two-layer GNN produces context-enriched
\emph{routing embeddings}:
\begin{equation}
\mathbf{Z}=\mathrm{GNN}(\mathbf{H}^0,\mathcal{G}_s)
    \in\mathbb{R}^{M\times d},
\label{eq:gnn}
\end{equation}
where $\mathbf{z}_m$ denotes the $m$-th row of $\mathbf{Z}$ and
$\mathbf{h}^0_m$ the $m$-th row of $\mathbf{H}^0$.
$\mathbf{Z}$ is used \emph{only} to parameterise routing costs; experts
aggregate content features $\mathbf{H}^0$ in \S\ref{sec:aggregation}.

\subsubsection{Transport cost.}
Each expert $e$ has a learnable prototype
$\boldsymbol{\mu}_e\in\mathbb{R}^{d}$.
The region-to-expert cost is cosine dissimilarity between the routing
embedding and the prototype:
\begin{equation}
C_{m,e}=1-\cos(\mathbf{z}_m,\,\boldsymbol{\mu}_e).
\label{eq:cost}
\end{equation}
Let $\mathbf{C}\in\mathbb{R}^{M\times E}$ denote the cost matrix with
entries $C_{m,e}$.

\subsubsection{Entropic OT with capacity marginals.}
We set region supply masses $\mathbf{r}\in\Delta_M$ with
$r_m\propto A_m$ (normalised) and expert capacities
$\mathbf{q}\in\Delta_E$ as $q_e=1/E$.
The routing plan is obtained by:
\begin{equation}
\Pi^{\star}=
\arg\min_{\Pi\ge 0}\;
\langle \Pi,\mathbf{C}\rangle
+\varepsilon\sum_{m,e}\Pi_{m,e}(\log\Pi_{m,e}-1)
\quad\text{s.t.}\quad
\Pi\mathbf{1}=\mathbf{r},\;\Pi^{\top}\mathbf{1}=\mathbf{q},
\label{eq:ot}
\end{equation}
solved by Sinkhorn scaling.
The key property is the column marginal
$\Pi^{\top}\mathbf{1}=\mathbf{q}$: it constrains per-slide expert load
in the dense routing plan $\Pi^{\star}$.
In contrast, softmax dispatch normalises locally per region and imposes
no constraint on aggregate utilisation; if one expert achieves slightly
higher affinity, it can absorb the majority of routing mass, a
self-reinforcing dynamic that leads to routing
collapse~\cite{fedus2022switch}.
ROAM enforces balanced utilisation without auxiliary load-balancing
losses and their sensitive
hyperparameters~\cite{wang2024auxiliary}.

\subsection{Graph-Regularised Sinkhorn Iterations}
\label{sec:graph_sink}

Capacity-constrained OT balances expert load but is agnostic to spatial
layout: adjacent regions in the same tumour nest may split across
experts.
To promote spatially coherent assignments, we interleave a graph
diffusion step within the Sinkhorn iterations.
After each row--column projection pair, we smooth the log-transport plan
over $\mathcal{G}_s$:
\begin{equation}
\log\Pi_{m,:} \leftarrow
    (1-\lambda_s)\,\log\Pi_{m,:}
    + \lambda_s \sum_{n\in\mathcal{N}(m)} w_{mn}\,\log\Pi_{n,:},
\label{eq:diffusion}
\end{equation}
where $\mathcal{N}(m)$ denotes the spatial neighbours of region $m$
(excluding $m$ itself),
$w_{mn}\propto\exp(-\|\mathbf{c}_m-\mathbf{c}_n\|_2^2/\tau)$ with
$\tau>0$ are heat-kernel edge weights normalised so that
$\sum_{n\in\mathcal{N}(m)}w_{mn}=1$, and
$\lambda_s\!\in\![0,1]$ controls smoothing strength.
Sinkhorn projections are then reapplied to restore marginal feasibility;
the interleaving repeats for $T$ iterations.
This is not a modified OT objective but a \emph{routing regulariser}
that biases iterates toward spatially smooth plans while preserving the
capacity marginals through re-projection.
Setting $\lambda_s{=}0$ recovers standard Sinkhorn, isolating the
effect of spatial regularisation in ablations (\S\ref{sec:ablation}).

\subsection{Sparse Dispatch and Expert Aggregation}
\label{sec:aggregation}

\subsubsection{Top-$k$ sparsification.}
We retain the $k$ largest entries per row of $\Pi^{\star}$ and rescale
them to preserve the row marginal $r_m$:
\begin{equation}
\gamma_{m,e}=
\begin{cases}
r_m\,\Pi^{\star}_{m,e}\Big/\!\sum_{e'\in\mathrm{top\text{-}k}(m)}
    \Pi^{\star}_{m,e'}
    & \text{if } e\in\mathrm{top\text{-}k}(m),\\[4pt]
0 & \text{otherwise},
\end{cases}
\label{eq:topk}
\end{equation}
where $\mathrm{top\text{-}k}(m)$ denotes the indices of the $k$ largest
entries in $\Pi^\star_{m,:}$.
This yields a sparse dispatch matrix
$\boldsymbol{\gamma}\in\mathbb{R}^{M\times E}_{+}$ with at most $k$
nonzero entries per row.
Expert $e$'s support set is
$\mathcal{S}_e=\{m \mid \gamma_{m,e}>0\}$.
Capacity balance holds exactly for $\Pi^{\star}$ and approximately after
top-$k$ sparsification.

\subsubsection{Per-expert gated attention pooling.}
Each expert $e$ applies an independent gated-attention
pooler~\cite{ilse2018abmil} over its support $\mathcal{S}_e$, operating
on content features $\mathbf{H}^0$:
\begin{equation}
s_e(m)=\mathbf{w}_e^{\top}\!
\bigl[\tanh(\mathbf{V}_e\mathbf{h}^0_m)
    \odot\sigma(\mathbf{U}_e\mathbf{h}^0_m)\bigr],
\quad
\beta^{(e)}_m=\softmax_{m\in\mathcal{S}_e}\!\bigl(s_e(m)\bigr),
\label{eq:expert_attn}
\end{equation}
\begin{equation}
\mathbf{o}_e=\sum_{m\in\mathcal{S}_e}\beta^{(e)}_m\,\mathbf{h}^0_m
    \in\mathbb{R}^{d}.
\label{eq:expert_embed}
\end{equation}

\subsubsection{Expert fusion and slide prediction.}
Expert embeddings are fused via gated attention and classified:
\begin{equation}
g_e=\softmax_{e}\!\bigl(\psi(\mathbf{o}_e)\bigr),
\quad
\mathbf{o}_{\mathrm{slide}}=\sum_{e=1}^{E} g_e\,\mathbf{o}_e,
\quad
\hat{\mathbf{y}}=\mathrm{MLP}\!\bigl(\mathbf{o}_{\mathrm{slide}}\bigr),
\label{eq:cls_head}
\end{equation}
where $\psi$ is a lightweight gating MLP.
The full pipeline is end-to-end differentiable and trained with
cross-entropy loss.

\section{Experimental Setup and Baselines}
\label{sec:setup}
\subsubsection{Datasets}

We evaluate four WSI tasks. For The Cancer Genome Atlas (TCGA) cohorts~\cite{tcga}, we use patient-stratified 5-fold cross-validation: (i) \textbf{non-small cell lung cancer (NSCLC) histology}: lung adenocarcinoma (LUAD) vs.\ lung squamous cell carcinoma (LUSC): 1{,}043 TCGA slides, with 2{,}206 Clinical Proteomic Tumor Analysis Consortium (CPTAC) slides~\cite{cptac} for external testing; (ii) \textbf{breast cancer (BRCA) subtype}: invasive ductal carcinoma (IDC) vs.\ non-IDC: 1{,}126 slides; (iii) \textbf{colorectal cancer (CRC) subtype}: colon adenocarcinoma (COAD) vs.\ rectum adenocarcinoma (READ): 600 slides; and (iv) \textbf{prostate grading} on the Prostate cANcer graDe Assessment (PANDA) dataset~\cite{panda}: 10{,}615 slides with six-class International Society of Urological Pathology (ISUP) grades 0--5.

\subsubsection{Baselines and Metrics}

We compare to MeanPool, MaxPool, ABMIL~\cite{ilse2018abmil}, CLAM-SB/MB~\cite{lu2021clam}, DSMIL~\cite{li2021dsmil}, TransMIL~\cite{shao2021transmil}, ILRA~\cite{xiang2023ilra}, and MoE/OT baselines MAMMOTH-ABMIL, MAMMOTH-TransM~\cite{shao2026mammoth}, PAMoE~\cite{wu2025pamoe}, and OTSurv~\cite{ren2025otsurv}.
All methods use identical frozen patch embeddings (UNI2-h~\cite{chen2024uni}) and the same training protocol unless stated otherwise. We report slide-level AUC for binary tasks (NSCLC, BRCA, CRC) and QWK for six-class PANDA.

\subsubsection{Implementations}
Models are trained with AdamW, cosine decay with 5-epoch warmup, cross-entropy, batch size 1, dropout 0.25, gradient clipping (max norm 1.0), up to 200 epochs with early stopping (patience 20).
Default lr $2{\times}10^{-4}$, wd $1{\times}10^{-5}$; ROAM uses lr $5{\times}10^{-4}$, wd $1{\times}10^{-4}$.
Bags exceeding 4{,}096 patches are randomly subsampled during training; full bags are used at evaluation.
ROAM uses projection dimension $d{=}512$, $M{=}256$ regions, region-graph $k_{\mathrm{NN}}{=}8$, $E{=}8$ experts with top-$2$ dispatch, 2-layer GraphSAGE~\cite{graphsage} for routing, OT regularisation $\varepsilon{=}0.1$, $T{=}20$ Sinkhorn iterations, and graph smoothing strength $\lambda_s{=}0.3$ applied for 3 smoothing steps within the unrolled routing procedure; expert pooling uses gated attention with $d_{\mathrm{attn}}{=}64$.
All experiments run on a single NVIDIA A100 40\,GB.

\section{Results and Analysis}
\begin{table*}[t]
\centering
\scriptsize
\caption{\textbf{WSI benchmark results with UNI2 patch encoders} (mean$\pm$std over 5-folds).
For \textbf{NSCLC (LUAD/LUSC)}, we report internal AUC on TCGA (\emph{int.}) and external AUC on CPTAC (\emph{ext.}); for \textbf{BRCA} and \textbf{CRC} we report internal AUC; for \textbf{PANDA} we report internal QWK. Best per column in \textbf{bold}; second best \underline{underlined}.}
\label{tab:main_uni2h}
\resizebox{\textwidth}{!}{%
\begin{tabular}{l | l c c c c c}
\toprule
\multirow{2}{*}{\textbf{Enc.}} &
\multirow{2}{*}{\textbf{Method}} &
\multicolumn{2}{c}{\textbf{NSCLC (LUAD/LUSC)}} &
{\textbf{BRCA}} &
{\textbf{CRC}} &
{\textbf{PANDA}} \\
\cmidrule(lr){3-4}
& &
\emph{int.} AUC$\uparrow$ &
\emph{ext.} AUC$\uparrow$ &
\emph{int.} AUC$\uparrow$ &
\emph{int.} AUC$\uparrow$ &
\emph{int.} QWK$\uparrow$ \\
\midrule
\multirow{13}{*}{\textbf{UNI2-h}}
& OTSurv~\cite{ren2025otsurv}                 & 0.964$\pm$0.014 & 0.786$\pm$0.021 & 0.884$\pm$0.017 & 0.661$\pm$0.055 & 0.873$\pm$0.019 \\
& MeanPool                                    & 0.965$\pm$0.013 & 0.800$\pm$0.018 & 0.881$\pm$0.023 & 0.671$\pm$0.046 & 0.893$\pm$0.012 \\
& MaxPool                                     & 0.964$\pm$0.017 & 0.823$\pm$0.021 & 0.867$\pm$0.023 & 0.636$\pm$0.046 & 0.811$\pm$0.015 \\
& ILRA~\cite{xiang2023ilra}                   & 0.974$\pm$0.012 & 0.827$\pm$0.023 & 0.898$\pm$0.021 & \textbf{0.702$\pm$0.022} & \underline{0.916$\pm$0.005} \\
& CLAM-MB~\cite{lu2021clam}                   & 0.969$\pm$0.020 & 0.831$\pm$0.017 & 0.901$\pm$0.021 & 0.697$\pm$0.050 & 0.911$\pm$0.009 \\
& ABMIL~\cite{ilse2018abmil}                  & 0.974$\pm$0.012 & 0.836$\pm$0.002 & 0.903$\pm$0.020 & 0.693$\pm$0.026 & 0.913$\pm$0.012 \\
& TransMIL~\cite{shao2021transmil}            & 0.974$\pm$0.015 & 0.836$\pm$0.023 & \underline{0.905$\pm$0.018} & 0.677$\pm$0.036 & 0.884$\pm$0.008 \\
& MAMMOTH-TransM~\cite{shao2026mammoth}       & 0.972$\pm$0.017 & 0.837$\pm$0.022 & 0.897$\pm$0.015 & 0.699$\pm$0.039 & 0.893$\pm$0.003 \\
& PAMoE~\cite{wu2025pamoe}                    & \underline{0.975$\pm$0.016} & 0.837$\pm$0.053 & 0.901$\pm$0.018 & 0.696$\pm$0.028 & 0.885$\pm$0.007 \\
& DSMIL~\cite{li2021dsmil}                    & 0.969$\pm$0.018 & \underline{0.844$\pm$0.016} & 0.897$\pm$0.014 & 0.687$\pm$0.032 & 0.873$\pm$0.003 \\
& MAMMOTH-ABMIL~\cite{shao2026mammoth}        & 0.972$\pm$0.016 & 0.842$\pm$0.028 & 0.897$\pm$0.014 & 0.696$\pm$0.039 & 0.892$\pm$0.005 \\
& CLAM-SB~\cite{lu2021clam}                   & 0.972$\pm$0.018 & 0.842$\pm$0.018 & 0.904$\pm$0.021 & 0.697$\pm$0.022 & 0.915$\pm$0.005 \\
& \hl{\textbf{ROAM (ours)}}                   & \textbf{\hl{0.976$\pm$0.015}} & \textbf{\hl{0.845$\pm$0.019}} & \textbf{\hl{0.905$\pm$0.014}} & \underline{\hl{0.699$\pm$0.030}} & \textbf{\hl{0.917$\pm$0.003}} \\

\bottomrule
\end{tabular}%
}
\end{table*}
\subsubsection{Quantitative Analysis}
Table~\ref{tab:main_uni2h} reports a controlled comparison
under frozen UNI2-h embeddings and shared patient-stratified splits.
In-domain TCGA performance is near-saturated (AUC within
${\sim}1$\,pp), so the informative setting is NSCLC external
generalisation (TCGA$\rightarrow$CPTAC), where ROAM achieves AUC
$0.845{\pm}0.019$, matching DSMIL ($0.844{\pm}0.016$) while
substantially outperforming OT-based aggregation without spatial
modelling (OTSurv, $0.786{\pm}0.021$).

MoE baselines PAMoE and MAMMOTH-ABMIL reach comparable external
means ($0.837$, $0.842$) but with higher fold variance
($\pm 0.053$ and $\pm 0.028$ vs.\ ROAM's $\pm 0.019$), and
PAMoE does not improve over its single-pathway counterpart
CLAM-SB ($0.842$ ext.), supporting the premise that experts
without load control do not reliably help.

On PANDA (10{,}615 slides, 6 classes), ROAM achieves the highest
QWK ($0.917{\pm}0.003$), suggesting capacity-balanced routing
scales to larger multi-class settings.

\begin{figure}[t]
\centering
\includegraphics[width=0.98\textwidth]{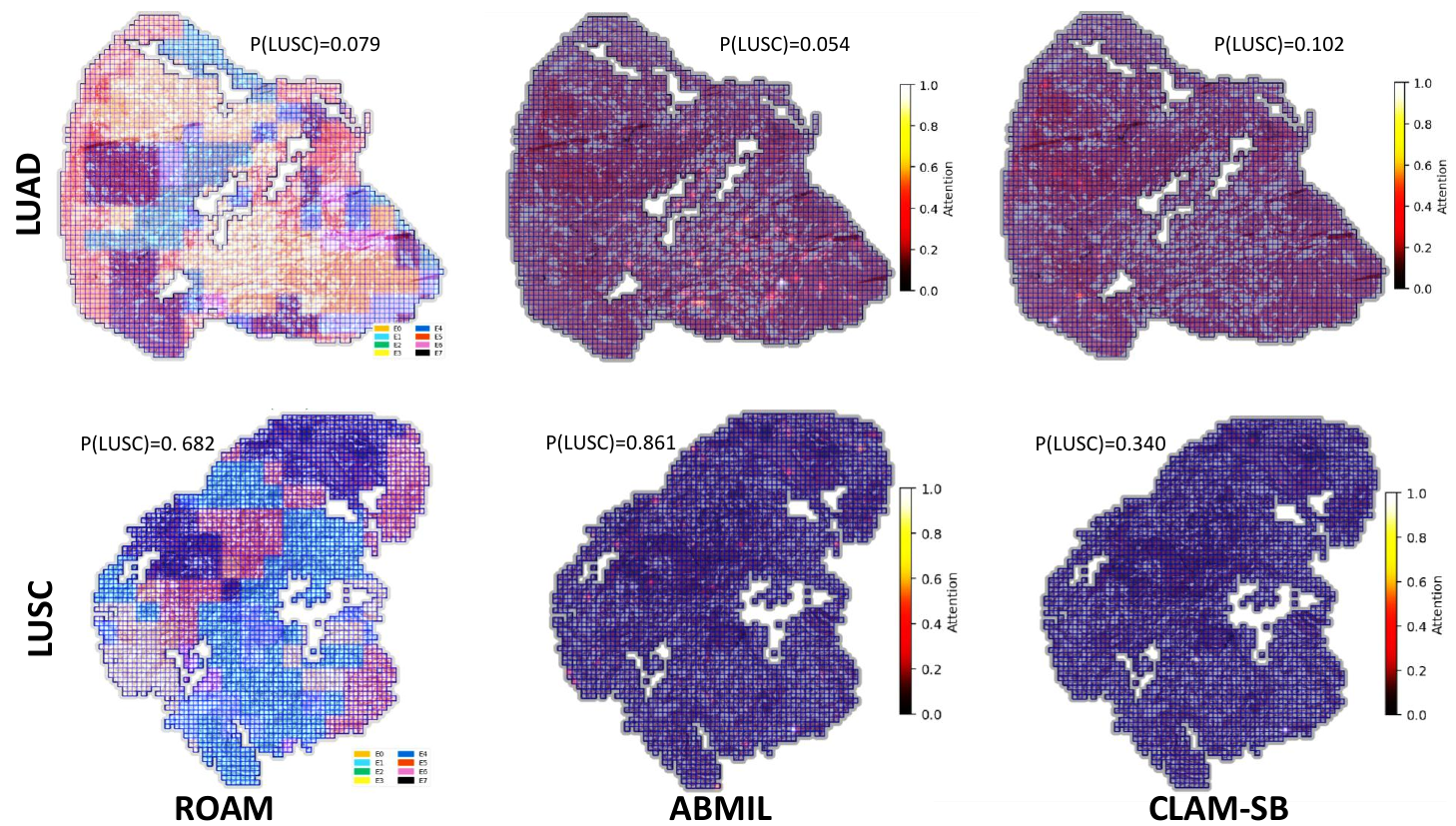}
\caption{Qualitative routing visualisation on CPTAC (NSCLC).
Two CPTAC external-test slides (top: LUAD; bottom: LUSC); numbers denote $P(\mathrm{LUSC})$.
Left: ROAM dominant expert per region (territories). Middle/Right: ABMIL and CLAM-SB attention heatmaps (normalised for visualisation).}
\label{fig:routing}
\end{figure}
\begin{table}[t]
\centering
\small
\setlength{\tabcolsep}{4pt}
\renewcommand{\arraystretch}{1.12}
\caption{\textbf{Ablation study on NSCLC (LUAD vs.\ LUSC) using UNI2-h embeddings} (mean$\pm$std over 5 folds).
Internal evaluation is on TCGA (\emph{int.}); external evaluation is on CPTAC (\emph{ext.}).}
\label{tab:ablation_nsclc_uni2h}
\begin{tabular}{@{}lccc@{}}
\toprule
   &\textbf{Configurations} & \emph{int.} AUC$\uparrow$ & \emph{ext.} AUC$\uparrow$ \\
\midrule
\addlinespace[1pt]
\textbf{I }  &\quad w/o routing GNN                                 & 0.971$\pm$0.022 & 0.836$\pm$0.023 \\
\textbf{II  } &\quad w/o graph regularisation                       & 0.966$\pm$0.019 & 0.813$\pm$0.035 \\
\textbf{III }  &\quad softmax routing (no capacity constraint)      & 0.964$\pm$0.029 & 0.809$\pm$0.045 \\
\textbf{IV }  &\quad w/o OT-guided pooling (no OT modulation)       & 0.966$\pm$0.026 & 0.825$\pm$0.015 \\
\textbf{V }  &\quad OT-guided pooling (detach routing weights)      & 0.970$\pm$0.019 & 0.820$\pm$0.055 \\
\midrule
\multicolumn{2}{c}{\textbf{ROAM}    }                                & \textbf{0.976$\pm$0.015} & \textbf{0.845$\pm$0.019} \\
\bottomrule
\end{tabular}
\end{table}
\subsubsection{Qualitative routing visualisation.}
Fig.~\ref{fig:routing} compares ROAM expert routing maps with
ABMIL and CLAM-SB attention heatmaps on two CPTAC slides.
These encode different quantities, discrete expert assignment
vs.\ continuous attention weight, so the comparison illustrates
routing \emph{structure}, not prediction quality.
ROAM produces spatially contiguous expert territories with all
eight experts receiving visible mass, consistent with the capacity
marginal $\Pi^{\top}\mathbf{1}{=}\mathbf{q}$
(\S\ref{sec:ot}) and graph smoothing (\S\ref{sec:graph_sink}).

\subsubsection{Ablation Studies}
\label{sec:ablation}
Table~\ref{tab:ablation_nsclc_uni2h} ablates ROAM on NSCLC, where the external CPTAC column is most discriminative.
Removing either capacity-constrained OT routing (III, $0.809$) or graph regularisation (II, $0.813$) degrades external performance relative to ROAM ($0.845$), indicating that both capacity control (\S\ref{sec:ot}) and spatial coherence (\S\ref{sec:graph_sink}) contribute to generalisation.
Detaching routing gradients (V) yields the largest fold variance ($\pm0.055$), suggesting that end-to-end supervision is important for stable routing in our setting.




\section{Conclusion}
We presented \textbf{ROAM}, a spatially aware MoE aggregator that routes region tokens to experts via capacity-constrained entropic optimal transport with a graph-based smoothing regulariser within unrolled Sinkhorn updates. Across four WSI benchmarks with frozen foundation-model embeddings, ROAM achieves competitive performance against strong MIL and MoE baselines; on NSCLC external generalisation (TCGA$\rightarrow$CPTAC) it achieves AUC $0.845{\pm}0.019$ and exhibits lower fold-to-fold variance than softmax-routed MoE baselines in our setting, consistent with the hypothesis that explicit capacity marginals stabilise routing under domain shift. Limitations include reliance on frozen encoders and fixed grid-based region tokenisation; adaptive regionisation and end-to-end fine-tuning are promising directions for future work.

\begin{credits}
\subsubsection{\discintname}
The authors have no competing interests to declare that are relevant to the content of this article.
\end{credits}


\bibliographystyle{splncs04}
\bibliography{references}

@inproceedings{ilse2018abmil,
  author    = {Ilse, Maximilian and Tomczak, Jakub M. and Welling, Max},
  title     = {Attention-based Deep Multiple Instance Learning},
  booktitle = {Proceedings of the 35th International Conference on Machine Learning (ICML)},
  volume    = {80},
  pages     = {2127--2136},
  publisher = {PMLR},
  year      = {2018},
}

@article{lu2021clam,
  author  = {Lu, Ming Y. and Williamson, Drew F. K. and Chen, Tiffany Y. and
             Chen, Richard J. and Barbieri, Matteo and Mahmood, Faisal},
  title   = {Data-efficient and weakly supervised computational pathology on
             whole-slide images},
  journal = {Nature Biomedical Engineering},
  volume  = {5},
  pages   = {555--570},
  year    = {2021},
}

@inproceedings{shao2021transmil,
  author    = {Shao, Zhuchen and Bian, Hao and Chen, Yang and Wang, Yifeng and
               Zhang, Jian and Ji, Xiangyang and Zhang, Yongbing},
  title     = {{TransMIL}: Transformer based Correlated Multiple Instance
               Learning for Whole Slide Image Classification},
  booktitle = {Advances in Neural Information Processing Systems (NeurIPS)},
  volume    = {34},
  year      = {2021},
}

@inproceedings{li2021dsmil,
  author    = {Li, Bin and Li, Yin and Eliceiri, Kevin W.},
  title     = {Dual-stream Multiple Instance Learning Network for Whole Slide
               Image Classification with Self-supervised Contrastive Learning},
  booktitle = {IEEE/CVF Conference on Computer Vision and Pattern Recognition (CVPR)},
  year      = {2021},
}

@inproceedings{xiang2023ilra,
  author    = {Xiang, Jinxi and Wang, Xiyue and Zhang, Jun and Yang, Sen and
               Han, Xiao and Yang, Wei},
  title     = {Exploring Low-Rank Property in Multiple Instance Learning for
               Whole Slide Image Classification},
  booktitle = {International Conference on Learning Representations (ICLR)},
  year      = {2023},
}

@inproceedings{shao2026mammoth,
  author    = {Shao, Daniel and Runevic, Joel and Chen, Richard J. and
               Williamson, Drew F. K. and Kim, Ahrong and Song, Andrew H. and
               Mahmood, Faisal},
  title     = {Mixture of Mini Experts: Overcoming the Linear Layer Bottleneck
               in Multiple Instance Learning},
  booktitle = {International Conference on Learning Representations (ICLR)},
  year      = {2026},
}

@inproceedings{wu2025pamoe,
  author    = {Wu, Junxian and Chen, Minheng and Ke, Xinyi and Xun, Tianwang and
               Jiang, Xiaoming and Zhou, Hongyu and Shao, Lizhi and Kong, Youyong},
  title     = {Learning Heterogeneous Tissues with Mixture of Experts for
               Gigapixel Whole Slide Images},
  booktitle = {IEEE/CVF Conference on Computer Vision and Pattern Recognition (CVPR)},
  year      = {2025},
}

@article{nguyen2025graphoftokens,
  author  = {Nguyen, Tam and Tran, Ngoc N. and Nguyen, Khai and Baraniuk, Richard G.},
  title   = {Improving Routing in Sparse Mixture of Experts with Graph of Tokens},
  journal = {arXiv preprint arXiv:2505.00792},
  year    = {2025},
}

@inproceedings{cuturi2013sinkhorn,
  author    = {Cuturi, Marco},
  title     = {Sinkhorn Distances: Lightspeed Computation of Optimal
               Transportation Distances},
  booktitle = {Advances in Neural Information Processing Systems (NeurIPS)},
  volume    = {26},
  pages     = {2292--2300},
  year      = {2013},
}

@inproceedings{ren2025otsurv,
  author    = {Ren, Qin and Wang, Yifan and Fang, Ruogu and Ling, Haibin and
               You, Chenyu},
  title     = {{OTSurv}: A Novel Multiple Instance Learning Framework for
               Survival Prediction with Heterogeneity-aware Optimal Transport},
  booktitle = {Medical Image Computing and Computer Assisted Intervention (MICCAI)},
  year      = {2025},
}

@article{chen2024uni,
  author  = {Chen, Richard J. and Ding, Tong and Lu, Ming Y. and
             Williamson, Drew F. K. and Jaume, Guillaume and Song, Andrew H. and
             Chen, Bowen and Zhang, Andrew and Shao, Daniel and Shaban, Muhammad and
             Williams, Mane and Oldenburg, Lukas and Weishaupt, Luca L. and
             Wang, Judy J. and Vaidya, Anurag and Le, Long Phi and Gerber, Georg and
             Sahai, Sharifa and Williams, Walt and Mahmood, Faisal},
  title   = {Towards a general-purpose foundation model for computational pathology},
  journal = {Nature Medicine},
  volume  = {30},
  pages   = {850--862},
  year    = {2024},
}

@article{fedus2022switch,
  title   = {Switch Transformers: Scaling to Trillion Parameter Models with Simple and Efficient Sparsity},
  author  = {Fedus, William and Zoph, Barret and Shazeer, Noam},
  journal = {Journal of Machine Learning Research},
  volume  = {23},
  pages   = {1--40},
  year    = {2022}
}

@book{villani2009optimal,
  title={Optimal transport: old and new},
  author={Villani, C{\'e}dric and others},
  volume={338},
  year={2009},
  publisher={Springer}
}

@article{knight2008sinkhorn,
  title={The Sinkhorn--Knopp algorithm: convergence and applications},
  author={Knight, Philip A},
  journal={SIAM Journal on Matrix Analysis and Applications},
  year={2008},
  publisher={SIAM}
}

@article{khamis2024scalable,
  title={Scalable Optimal Transport Methods in Machine Learning: A Contemporary Survey},
  author={Khamis, Abdelwahed and Tsuchida, Russell and Tarek, Mohamed and Rolland, Vivien and Petersson, Lars},
  journal={IEEE Transactions on Pattern Analysis and Machine Intelligence},
  year={2024}
}

@article{graphsage,
  title={Inductive representation learning on large graphs},
  author={Hamilton, Will and Ying, Zhitao and Leskovec, Jure},
  journal={Advances in neural information processing systems},
  volume={30},
  year={2017}
}

@inproceedings{wang2024auxiliary,
  title={Auxiliary-Loss-Free Load Balancing Strategy for Mixture-of-Experts},
  author={Wang, Lean and Huang, Huazuo and Wu, Shuming and Ma, Shuming and Wei, Furu},
  booktitle={ICLR},
  year={2025}
}

@inproceedings{fang2024sammil,
  title={{SAM-MIL}: A Spatial Contextual Aware Multiple Instance Learning Approach for Whole Slide Image Classification},
  author={Fang, Heng and Huang, Sheng and Tang, Wenhao and Huangfu, Luwen and Liu, Bo},
  booktitle={ACM Multimedia},
  year={2024}
}

@inproceedings{zhou2022expertchoice,
  title     = {Mixture-of-Experts with Expert Choice Routing},
  author    = {Zhou, Yanqi and Lei, Tao and Liu, Hanxiao and Du, Nan and Huang, Yanping and Zhao, Vincent and Dai, Andrew and Chen, Zhifeng and Le, Quoc and Laudon, James},
  booktitle = {NeurIPS},
  volume    = {35},
  year      = {2022}
}

@article{gigapath,
  title={A whole-slide foundation model for digital pathology from real-world data},
  author={Xu, Hanwen and Usuyama, Naoto and Bagga, Jaspreet and Zhang, Sheng and Rao, Rajesh and Naumann, Tristan and Wong, Cliff and Gero, Zelalem and Gonz{\'a}lez, Javier and Gu, Yu and others},
  journal={Nature},
  volume={630},
  number={8015},
  pages={181--188},
  year={2024},
  publisher={Nature Publishing Group UK London}
}

@article{campanella2019clinical,
  author  = {Campanella, Gabriele and Hanna, Matthew G. and Geneslaw, Luke and
             Miraflor, Allen and Silva, Vitor Werneck Krauss and Busam, Klaus J. and
             Brogi, Edi and Reuter, Victor E. and Klimstra, David S. and
             Fuchs, Thomas J.},
  title   = {Clinical-grade computational pathology using weakly supervised deep
             learning on whole slide images},
  journal = {Nature Medicine},
  volume  = {25},
  number  = {8},
  pages   = {1301--1309},
  year    = {2019},
}

@article{tcga,
  title={The cancer genome atlas pan-cancer analysis project},
  author={Weinstein, John N and Collisson, Eric A and Mills, Gordon B and Shaw, Kenna R and Ozenberger, Brad A and Ellrott, Kyle and Shmulevich, Ilya and Sander, Chris and Stuart, Joshua M},
  journal={Nature genetics},
  volume={45},
  number={10},
  pages={1113--1120},
  year={2013},
  publisher={Nature Publishing Group}
}

@article{cptac,
  title = {{The CPTAC Data Portal: A Resource for Cancer Proteomics Research}},
  author = {Edwards, Nathan J. and Oberti, Mauricio and Thangudu, Ratna R. and Cai, Shuang and McGarvey, Peter B. and Jacob, Shine and Madhavan, Subha and Ketchum, Karen A.},
  journal = {Journal of Proteome Research},
  volume = {14},
  number = {6},
  pages = {2707--2713},
  year = {2015},
  month = jun,
  issn = {1535-3893},
  publisher = {American Chemical Society},
}

@article{panda,
  title={{Artificial intelligence for diagnosis and Gleason grading of prostate cancer: the PANDA challenge}},
  author={Bulten, Wouter and Kartasalo, Kimmo and Chen, Po-Hsuan Cameron and Str{\"o}m, Peter and Pinckaers, Hans and Nagpal, Kunal and Cai, Yuannan and Steiner, David F and Van Boven, Hester and Vink, Robert and others},
  journal={Nature medicine},
  volume={28},
  number={1},
  pages={154--163},
  year={2022},
  publisher={Nature Publishing Group US New York}
}

\end{document}